\documentclass{article}

\PassOptionsToPackage{round}{natbib}

\usepackage[final]{nips_2018}

\usepackage[utf8]{inputenc} 
\usepackage[T1]{fontenc}    
\usepackage{url}            
\usepackage{booktabs}       
\usepackage{amsfonts}       
\usepackage{nicefrac}       
\usepackage{microtype}      
\usepackage{graphicx}  
\usepackage{subcaption}
\usepackage{listings,newtxtt}
\usepackage[colorlinks = true,
            linkcolor = blue,
            urlcolor  = blue,
            citecolor = blue,
            anchorcolor = blue]{hyperref}

\usepackage[textsize=small,textwidth=\marginparwidth]{todonotes}

\begin{document}

\title{NAIL: A General Interactive Fiction Agent}

\author{
  Matthew Hausknecht\thanks{Equal contribution}\\
  \And
  Ricky Loynd\footnotemark[1]\\
  \And
  Greg Yang\\
  \And
  Adith Swaminathan\\
  \And
  Microsoft Research AI\\
  \texttt{\{mahauskn,riloynd,gregyang,adswamin\}@microsoft.com} \\
  \And
  Jason D. Williams\thanks{Work done while author was at Microsoft Research.}\\
  Apple\\
  \texttt{jdw@alumni.princeton.edu}\\
}

\maketitle
\begin{abstract}
Interactive Fiction (IF) games are complex textual decision making problems. This paper introduces NAIL, an autonomous agent for general parser-based IF games. NAIL won the 2018 Text Adventure AI Competition, where it was evaluated on twenty unseen games. This paper  describes the architecture, development, and insights underpinning NAIL's performance.\footnote{NAIL's source code  is available at~\href{https://github.com/Microsoft/nail_agent}{https://github.com/Microsoft/nail\_agent}.}
\end{abstract}

\section{Introduction}
\begin{figure}[htp]
\centering
\begin{subfigure}{.4\linewidth}
  \centering
  \includegraphics[width=\linewidth]{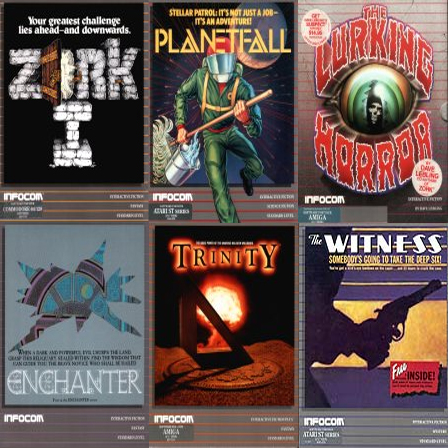}
  \label{fig:infocom_left}
\end{subfigure}%
\begin{subfigure}{.4\linewidth}
  \centering
  \includegraphics[width=1.02\linewidth]{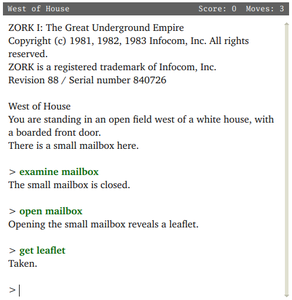}
  \label{fig:infocom_right}
\end{subfigure}
\vspace{-2em}
\caption{\textbf{Interactive fiction games} made Infocom the dominant computer game company of the early 1980s. Right: Transcript of Zork, with player actions in green.}
\label{fig:infocom}
\end{figure}

Interactive Fiction games are rich simulation environments where players issue text commands to control their character and progress through the narrative. Unlike modern video games, there are no graphics; text is the sole modality of interaction. Of the varieties of IF games, this paper is concerned with parser-based IF games, environments in which the player may issue arbitrary textual commands which the game's parser attempts to interpret and execute. Despite this daunting interface, parser-based IF games were highly successful in the early 1980s. Classic games like Zork (Figure \ref{fig:infocom}) and Hitchhiker's Guide to the Galaxy sparked the imaginations of many and created a community of IF enthusiasts who continue to play and create new games to this day.

As a testbed for autonomous agents, IF games promote natural language understanding, commonsense reasoning, problem solving, and planning. Formally, IF games can be understood as Partially Observable Markov Decision Processes (POMDPs), since they are governed by an underlying system but provide only partial observations of the underlying game state. Game playing agents need to be capable of generating valid actions that make sense in the context of the current story. These actions often take the form of 1) information gathering about the player's surroundings, 2) interaction with nearby objects or people, and 3) navigation between game locations. In return for solving puzzles, the player is rewarded with game score that quantifies advancement through the story.

IF games present several unique challenges to learning agents: First, IF games feature a combinatorial language-based action space. Learning agents have been extensively studied in both discrete and continuous action spaces, but not the repeated discrete space defining IF actions. Second, IF games expect the player to understand how to interact with everyday objects like doors and mailboxes. This form of commonsense reasoning is difficult for learning agents lacking embodiment. Finally, IF games are extremely partially observable - they commonly feature tens to hundreds of unique locations, and the player only receives the description of its current location. Humans often construct a maps to remember how to navigate between rooms and keep track of which objects are in each location. For these reasons, learning agents that excel in graphical video games aren't directly applicable to IF games. Instead, a new type of agent is needed. 

NAIL (Navigate, Acquire, Interact and Learn) is an autonomous agent designed to play arbitrary human-made IF games. NAIL competed in the 2018 Text-Based Adventure AI Competition~\citep{atkinson18}, where it was evaluated on twenty unknown IF games, with only one-thousand steps of interaction per game. To excel in this context, NAIL needed to be truly general, with the capability of encountering a new game and quickly accruing as much score as possible. For this reason, we designed NAIL with strong heuristics for exploring the game, interacting with objects, and building an internal representation of the game world. The remainder of this paper describes NAIL's architecture, development process, and innovations.

\section{Related Work}

NAIL draws inspiration from previous generations of IF game-playing agents:

\noindent \textbf{BYU`16 Agent}~\citep{fulda17} analyzes relationships in word-embedding space~\citep{mikolov13} to reason about which objects are possible to interact with and which verbs can likely be paired with each interactive object. Armed with this knowledge, the agent exhaustively enumerates possible actions. While effective, this approach is less efficient in terms of the number of in-game steps required. 

\noindent \textbf{Golovin}~\citep{kostka17} generates actions using a large set of predefined command patterns extracted through walkthroughs, tutorials, and decompilation of games. Additionally, it employs specialized command generators for common tasks such as fighting, collecting items, managing inventory, and exploring. An recurrent neural network is used to identify the objects most likely to be interactable in a scene's description.

\noindent \textbf{CARL}~\citep{atkinson18} was developed by the creators of the BYU Agent and uses skip-thought vectors~\citep{kiros15} as representations of the observation text. These vectors are classified as either containing location information or relating to the effects of an action. For those containing location information, commands are generated by extracting nouns from the text and using the BYU Agent's word embedding approach to find likely verbs. A hash table is used to keep track of visited states and actions to avoid repetition.

There is also a growing body of work studying reinforcement learning agents in textual environments: LSTM-DQN~\citep{narasimhan15} introduced the idea of building a representation for the current state by processing narrative text with a recurrent network. Building on this idea, LSTM-DRQN~\citep{yuan18} added a second level recurrence over states, which was demonstrated to give the agent an ability to reason over locations visited in the past. In the realm of choice-based games, \cite{he16,zelinka18} explored architectural choices for scoring different pre-defined actions. Another recent innovation included training a separate network to eliminate consideration of invalid actions ~\citep{haroush18}. Finally, \cite{ammanabrolu18}  demonstrated end-to-end learning using an architecture that represents game state as a knowledge graph.

\section{Architecture}
\begin{figure}[htp]
\centering
\includegraphics[width=.85\linewidth]{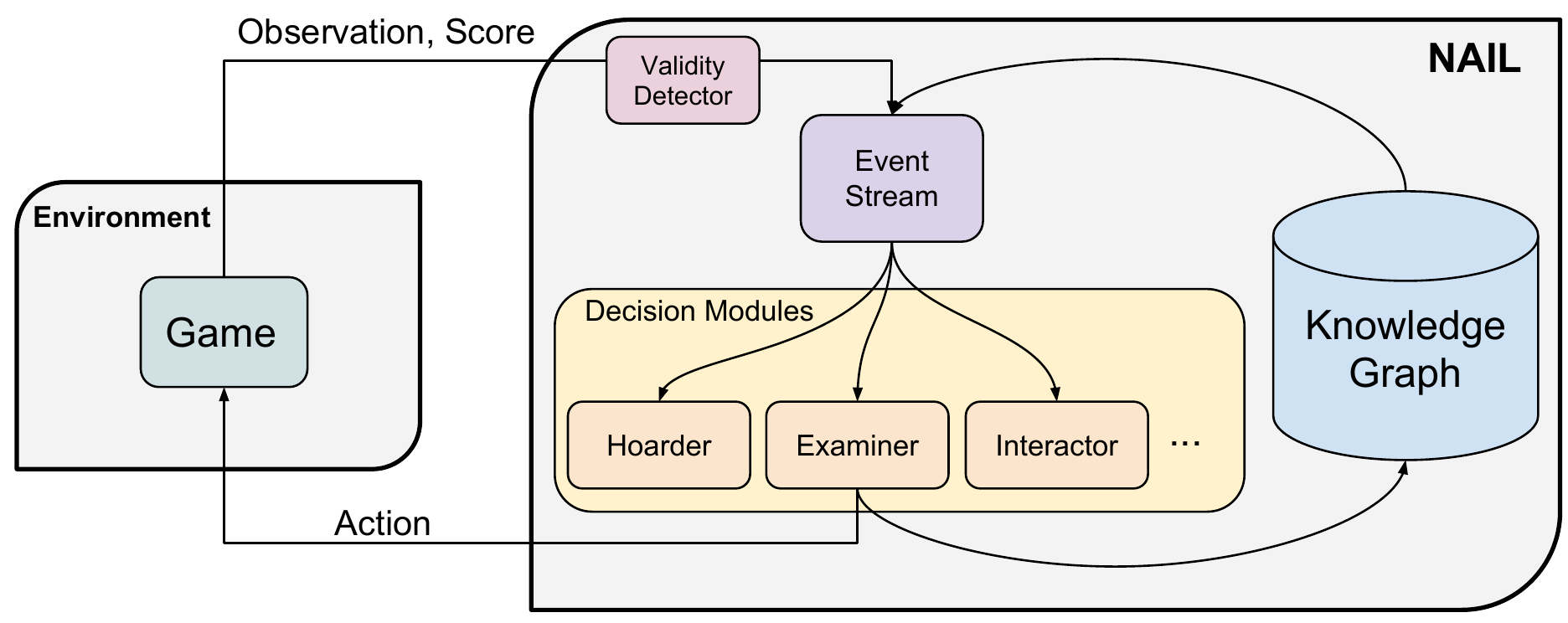}
\caption{\textbf{NAIL} consists of multiple Decision Modules, which are designed to perform specialized tasks. One decision module at a time may be active. The active module is responsible for generating actions and updating the knowledge graph with the effects of its actions. The \textit{knowledge graph} builds a structured representation of the objects, locations, and interactions observed in the game so far and is used by the decision modules to select actions.}
\label{fig:nail_arch}
\end{figure}

NAIL was designed with a modular architecture that decouples the decision making from the knowledge acquisition. At the highest level, NAIL utilizes a set of Decision Modules to interact with the game and accumulates information in its knowledge graph. Further separation is achieved by decomposing the overall decision making problem into specialized task-specific Decision Modules, which take control of the agent to perform a particular job when the right context arises. For example, the \textit{darkness} decision module activates in response to game narratives that include key phrases "pitch black" and "too dark too see," and attempts to provide light by issuing a "turn on" command. Separating decision making from knowledge acquisition enables NAIL's capabilities to be augmented by adding more decision modules while being able to inspect and debug the information accumulated in the knowledge graph. Figure \ref{fig:nail_arch} depicts NAIL's architecture, which we discuss in detail below.

\section{Validity Detector}
\label{sec:valid}
The game's responses to the agent's actions provide a continual stream of feedback. NAIL processes game responses to determine whether an action was 1) unrecognized by the game, 2) recognized but failed, or 3) recognized and successful. This information is crucial to determining how the Knowledge Graph is populated with Locations and Entities.


\begin{table}
\centering
\begin{tabular}{ll}
Response & $p(\textrm{success})$\\
\hline
I didn’t understand that sentence. & 0 \\
You can’t go that way. & 0 \\
You can’t use multiple objects with that verb. & .0008 \\
You try to push past, but vines block your way. & .0009 \\
I don’t know the word xyzzy. & .1145 \\
Even with a lamp, you would not chance these stairs in the darkness. & .6366 \\
The gentle tapping sounds again. & .9387 \\
Help! You hurtle through the cave opening! & .9835 \\
The grating opens. & .9998 \\ 
The cyclops seems somewhat agitated. & .9998 \\
\end{tabular}
\vspace{1em}
\caption{\textbf{Validity Detector} predicts if the game's response to an action indicates success or failure. Predictions are only concerned with whether or not the action was recognized and effected change in the game. Responses are classified as successful even if they indicate that the player has done something wrong or is likely to die.}
\label{tab:responses}
\end{table}

The Validity Detector, a FastText~\citep{joulin2017bag} classifier, predicts whether the game's response to an action indicates success or failure. The Validity Detector was trained on a manually created dataset of 1337 successful responses and 705 failed responses. Despite the small size of the training dataset, classifier performance was surprisingly good, as shown by the example predictions in Table \ref{tab:responses}. Many games use a common set of failure responses, simplifying the classification task. 

The Validity Detector underlies each of NAIL's core components: it is used by Decision Modules to decide when actions have succeeded and it serves as a gatekeeper to the knowledge graph by disallowing the creation of invalid Entities and Locations.

\section{Decision Modules}
NAIL is composed of many decision modules (Fig. \ref{fig:nail_arch}) which can be thought of as sub-policies designed to accomplish individual tasks. NAIL's core decision modules are the Examiner, Hoarder, Interactor, and Navigator. Respectively they are responsible for identifying relevant objects, acquiring objects, interacting with identified objects, and changing locations. Before delving into the specifics of these core modules, we first discuss the process for synchronizing control between modules.

\subsection{Choosing Between Decision Modules: Eagerness}
Each decision module has the ability to take control of the agent and generate textual actions. Since only a single decision module can have control over the agent at any given timestep, NAIL employs a winner-take-all competition in which decision modules report how eager they are to have control over the agent given the current context. Game context consists of various factors such as information contained within the game narrative and knowledge graph. Different modules are eager in different contexts: the Examiner module is eager to take control when a new location is discovered, while the Interactor is most eager when new objects are added to the knowledge graph. By convention, eagerness values are real numbers constrained to the range $(0, 1)$.

The most eager decision module generates actions until it relinquishes control. Currently there is no mechanism for decision modules to interrupt each other, as this would require the interrupted module to resume control in a different context than before the interrupt. Instead, modules are designed to accomplish many short tasks and relinquish control so as to provide more eager modules many chances to take control.

\subsection{Examiner Decision Module}
One of the fundamental steps in playing IF games is reading the current narrative and deciding which parts of the surroundings are capable of being interacted with. This is the job of the Examiner, a decision module that outputs a high eagerness upon visiting a new location\footnote{More precisely upon a new Location object being added to the knowledge graph.} and seeks to identify objects to add to the knowledge graph. Identification of objects proceeds in two steps: first, candidate noun phrases are extracted from the narrative text using Spacy~\citep{spacy2}. Second, "examine X" commands are issued for each of the candidate noun phrases. If the game's response to the examine command indicates the object is recognized (often in the form of a longer description of that object) the Examiner will add the object to the knowledge graph. Conversely if the response indicates the game doesn't recognize the object, it is not added to the knowledge graph and will not be interacted with in the future. The Validity Detector is used to decide which responses are valid.

\begin{figure}[htp]
\centering
\includegraphics[width=.6\linewidth]{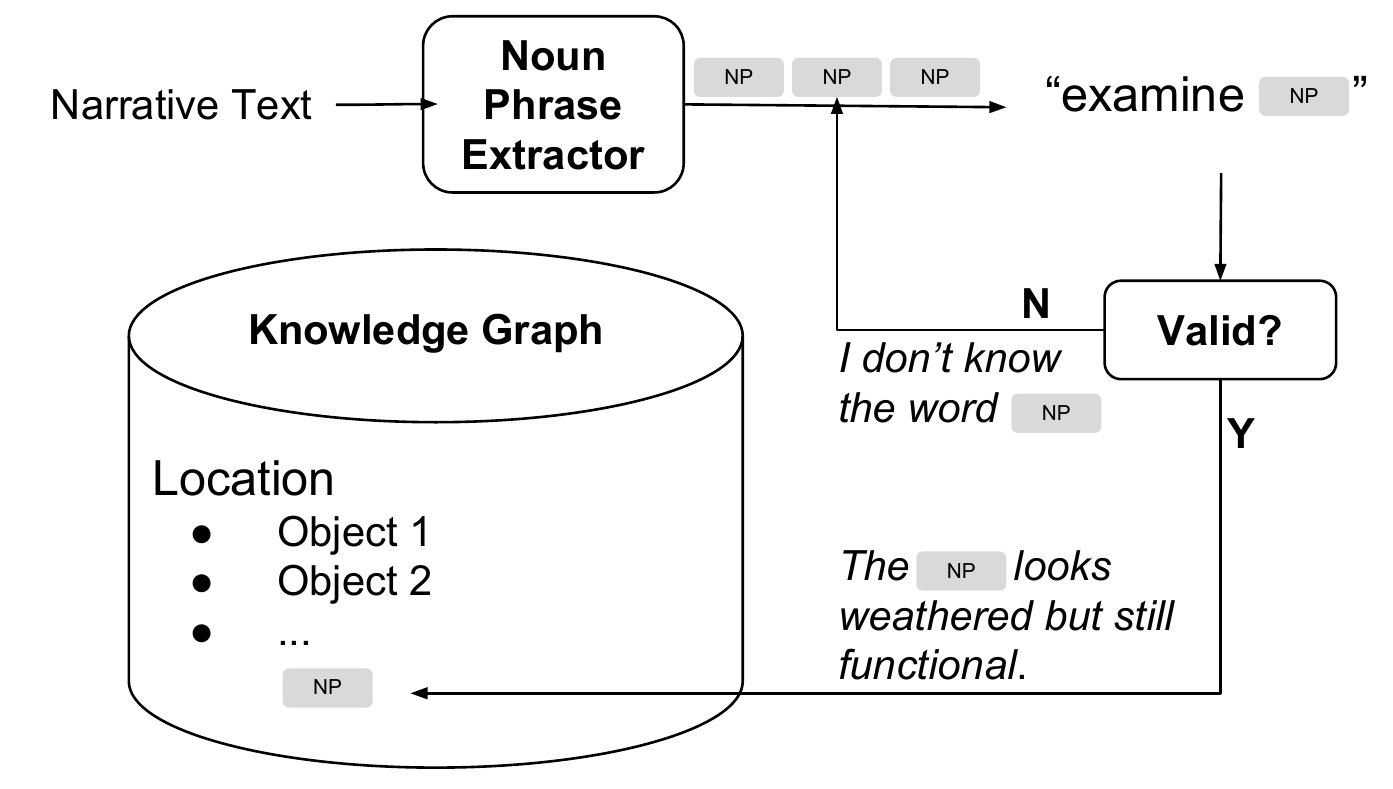}
\caption{\textbf{The Examiner} decision module is responsible for extracting game objects from narrative text and adding them to the knowledge graph. Subsequent modules like the Interactor use the objects in the knowledge graph to generate further actions.}
\label{fig:examiner}
\end{figure}

\subsection{Hoarder Decision Module}
Acquiring objects such as weapons, treasures, and consumables is a necessity in nearly all IF games. The Hoarder outputs the highest possible eagerness upon reaching a new location. After taking control it issues a single "take all" command. Many games support this shorthand and will reward the player by transporting all portable objects directly into the player's inventory.\footnote{For the games that don't support the take all command, objects may still be acquired individually using the Interactor decision module.} For example, in Hitchhiker's Guide to the Galaxy:

\texttt{> take all\\
telephone: You lunge for it, but the room spins nauseatingly away.\\
flathead screwdriver: It slips through your fumbling fingers and hits the carpet with a nerve-shattering bang.\\
toothbrush: You lunge for it, but the room spins nauseatingly away.\\
your gown: Luckily, this is large enough for you to get hold of. You notice something in the pocket.}

The response indicates four valid objects are present, of which we succeeded in acquiring only the gown. The Hoarder uses a custom parser to add all detected objects to the knowledge graph and the successfully acquired objects to the player's inventory. The Validity Detector decides whether a response indicates the item has been successfully acquired.

\subsection{Interactor Decision Module}
\begin{figure}[htp]
\centering
\includegraphics[width=.8\linewidth]{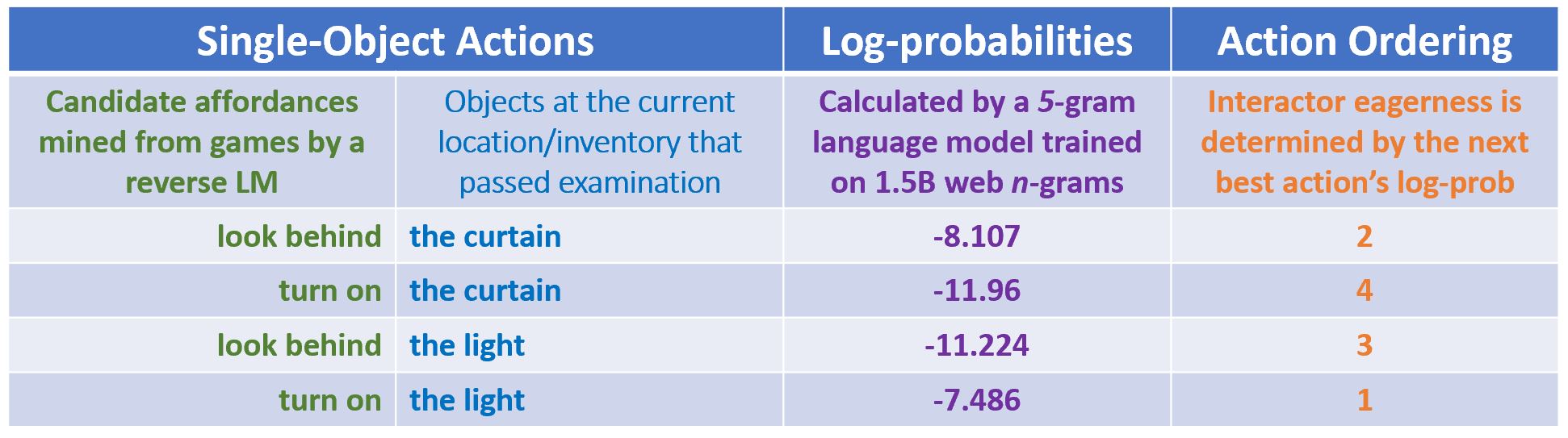}
\caption{\textbf{The Interactor} ranks candidate actions according to their log-probabilities. These log-probabilities are proportional to the Interactor's eagerness.}
\label{fig:interactor}
\end{figure}

A core challenge of IF games is deciding how to effectively interact with the objects at the current location. Exhaustive search is rarely tractable as even a restricted subset of the English language can produce a combinatorially large set of possible actions. To narrow this search, NAIL uses a \textit{5}-gram language model (LM), trained on 1.5 billion \textit{n}-grams collected from web page titles,\footnote{Although language used in web titles is definitely not the same as actions in IF games, in terms of both speed and perplexity, LMs trained on titles have been reported to outperform LMs trained on document body text on general information retrieval tasks~\citep{wang1}.} to impose a prior over generated actions. As shown in Figure \ref{fig:interactor}, this prior gives preference to common, sensible actions like "open the door" or "light the torch" over less reasonable actions like "light the door" or "open the torch." The Interactor works through the prioritized list of actions in descending order, executing a single action, judging whether it succeeded, and reporting an eagerness score proportional to how highly ranked the next action is.

In more detail, starting from the set of objects present in the player's current location, the Interactor generates an unranked list of candidate actions. These candidate actions are enumerated from two templates: a verb-object template and a verb-object-preposition-object template.\footnote{As discussed in Appendix \ref{sec:appendix_analysis}, the vast majority of actions taken by human players can be expressed by one of these two templates.} Verb-object commands use verbs selected from a manually curated list of 561 commonly recognized IF-game verbs. These verbs are paired with the objects present at the current location and inventory; the LM is then used to compute the joint probability of each generated verb-object pair. 

Verb-object-preposition-object commands are also created by iterating over objects $x, y$ present at the current location or in the player's inventory. Ten fixed templates (such as \textit{put x in y} and \textit{open x with y}) provide verbs and prepositions for each object pair. These candidate commands are then ranked along with the verb-object actions according to their LM-based joint probabilities.\footnote{LM probabilities for two-word actions are nearly always higher than four-word actions. Fortunately, as shown in Figure \ref{fig:action_hist}, humans also prefer shorter actions.}

\subsection{Navigator Decision Module}
The Navigator is invoked to move the player to a new location. Outputting a consistently low eagerness, the Navigator is intended to activate only after the Hoarder, Examiner, and Interactor have exhausted the useful actions at the current location. After gaining control, the Navigator applies one of the twelve canonical navigation actions: North, South, West, East, Northwest, Northeast, Southwest, Southeast, Up, Down, Enter, and Exit. To avoid repeating failed navigation actions, the Navigator keeps track of all previously attempted navigational actions from each location, and prefers actions that have either not yet been attempted or have succeeded in changing locations in the past. Additionally, since location descriptions frequently tell the player where exits are (e.g. "there is an open door to the west"), the Navigator performs a simple string match and prioritizes directional actions that are present in the description.

After attempting a navigational action the Navigator must decide whether or not the action succeeded and a new location was reached. To do this, it first checks the knowledge graph to see if there are any locations whose textual description matches the game's response. If no known locations are found, the Navigator issues a "look" command and compares the resulting description to the description of the previous location. High similarity between location descriptions indicates that the move action likely failed; low similarity indicates that a new location has likely been discovered.

A fuzzy string match~\citep{fuzzywuzzy} between location descriptions is necessary since many games randomly alter location descriptions. For example, in the forested locations of Zork 1, the game will occasionally append "You hear in the distance the chirping of a song bird" to the location's description. Similarly, dropped objects will be reported in a location's description.

Finally, to finish a location change, the Navigator updates the Knowledge Graph by adding a new location and connecting it to the previous location. This step results in an ad-hoc map of the game world. In future work, this map could be used to efficiently revisit previous locations or to guide future exploration.

\subsection{Specialized Decision Modules}
NAIL uses several highly specialized decision modules, mainly ensure the agent doesn't get stuck while playing a game.

\begin{itemize}
    \item Darkness: Emits a "turn on" action in response to observation text containing phrases "pitch black" or "too dark to see".
    \item Restart: Emits a "restart" action to restart the game in response to an observation containing "restart", "restore", and "quit" being observed.
    \item Yes-No: Randomly emits either "Yes" or "No" in response to common game prompts asking the player to answer a yes-no question.
    \item YouHaveTo: Attempts to take advantage of in-game hints by using a set of custom regular expressions to parse suggested actions from the game's response. For example: "You'll have to get out of bed first" will emit "get out of bed" as the next action.
    \item Idler: As a fallback when no other decision module is eager for control, the Idler randomly composes actions from hundreds of common verb phrases combined with nearby objects. This exhaustive exploration over possible actions sometimes produces combinations that can get the agent unstuck.
\end{itemize}

Due to the flexibility of NAIL's architecture it's quite easy to create new decision modules to handle different situations. 

\subsection{Decision Modules as Python Generators}
A key design choice was to implement decision modules as Python generators. Using Python's \texttt{yield} to generate actions allows a straight-line implementation of the logic within each DM. Consider the case of a \texttt{MorningRoutine} decision module. Using a generator, the implementation is straightforward:

\begin{lstlisting}[language=Python]
class MorningRoutine(DecisionModule):
    def take_control(self):
        obs = yield
        obs = yield `get out of bed'
        obs = yield `turn on light'
        obs = yield `brush teeth'
\end{lstlisting}

Without a generator the logic for the same DM becomes quite a bit more complex:

\begin{lstlisting}[language=Python]
class MorningRoutine(DecisionModule):
    def __init__(self):
        self.out_of_bed = False
        self.turned_on_light = False
        self.brushed_teeth = False

    def take_control(self, obs):
        if not self.out_of_bed:
            self.out_of_bed = True
            return `get out of bed'
        (*\bfseries elif not*) self.turned_on_light:
            self.turned_on_light = True
            (*\bfseries return*) `turn on light'
        elif not self.brushed_teeth:
            self.brushed_teeth = True
            return `brush teeth'
\end{lstlisting}

In other words, generators allow the agent to be written as if it has direct access to the environment's \texttt{step} function when, in reality, it is being invoked as a library and does not have direct access to the environment. This is commonly the case in competition settings.

\section{Knowledge Graph}
NAIL accumulates knowledge about the game world as the agent interacts with the game. Specifically, NAIL's knowledge graph keeps track of objects, past interactions, locations, connections between locations, object states, and unrecognized words. This information is used by decision modules to compute eagerness and generate actions. In turn, the decision modules modify the knowledge graph to reflect the consequences of their actions.

At the top level, the knowledge graph is organized as follows:

\begin{itemize}
\item \textbf{Current Location}: Player's current location, updated by the Navigator upon successful movement.
\item \textbf{Locations}: List of all discovered locations.
\item \textbf{Connection Graph}: Graph of connections between discovered locations. Updated by the Navigator upon successful movement.
\item \textbf{Inventory}: List of objects in the player's inventory. Updated by any decision module that issues take/drop commands.
\item \textbf{Unrecognized Words}: List of words not recognized by the game. NAIL avoids taking actions containing any unrecognized words. Updated after each action by matching the game's response against a custom list of unrecognized responses (e.g. "That's not a verb I recognise.").
\end{itemize}

Each Location contains the following information:
\begin{itemize}
    \item \textbf{Name}: The short name of the location (e.g. "West of House").
    \item \textbf{Description}: The full-length description of the location, as returned by a "look" command. Populated by the Navigator upon discovering the location.
    \item \textbf{Entities}: List of entities (interactive objects or characters) present at that location. Populated by the Examiner.
    \item \textbf{Action Record}: List of all actions NAIL has attempted at this location along with the game's response and NAIL's estimate of how likely the response indicates success. This information is optionally used by decision modules to avoid repeating actions that previously failed.
\end{itemize}

Finally, each Entity contains the following information:
\begin{itemize}
    \item \textbf{Names}: List of discovered names for this entity. Many games are flexible when referring to entities - e.g. the Brass Lantern in Zork may be alternatively referred to as "lantern" or even just "brass."
    \item \textbf{Description}: Long-form description of the entity - as given by "examine entity." Populated by Examiner.
    \item \textbf{Entities}: List of contained entities - e.g. in the case of a container such as a chest.
    \item \textbf{State}: Keeps track of a list of manually-defined states: Open/Closed, Locked/Unlocked, On/Off. Also keeps track of whether the item has been used - in the case of consumable items.
    \item \textbf{Attributes}: A manually-defined list of object attributes: Openable, Lockable, Switchable. Attributes inform which verbs are expected to succeed on a particular object.
\end{itemize}

Beyond its use to Decision Modules, the knowledge graph also provides an interpretable representation of NAIL's understanding of the game. By comparing the knowledge graph to the published map for well documented games like Zork, it was possible to track down bugs in NAIL's decision modules.

\section{Encoding Action Effects}
The core interaction for many games relies heavily on a small set of common actions - \textit{take, drop, turn on, push, pull, go north}, etc. Furthermore, the effects of these common actions are reasonably
general across games. For example, the \textit{take} action, if successful, will move an object from the player's current location to the player's inventory. 

For these common actions, the expected changes to the knowledge graph are manually implemented and associated with the action. This association allows the effects of the action to be implemented once, and subsequently used by many different decision modules. For non-common actions, we do not make any changes to the knowledge graph, aside from recording the action and its probability of success. In future work, it may be possible to learn the effects of uncommon actions.

\section{Text-Adventure Competition}
\label{sec:cig}
To meet the needs of generality and efficient exploration of unseen games, we developed and evaluated NAIL on a set of fifty-six IF games (full list in Table \ref{tab:scores-full}) using the \cite{jericho} Learning Environment. Jericho was a ideal learning environment because of its ability to introspect and provide ground truth knowledge of the game state. Our primary metric was normalized game score averaged over all games. However, due to the sparse rewards in most games, improvements to the NAIL agent often weren't reflected in game score. To address this problem, we created Dependencies, a fine-grained metric that quantifies progress towards the first point of score on each game. Specifically this metric manually defines the locations needed to be visited, the items that need to be acquired, entities that need to be detected, and the key actions that must be performed. The following snippet shows the Dependencies for the game Balances:

\begin{lstlisting}[language=Python]
analyzer.deps = [
    EntDep(['wooden furniture', 'furniture'], loc=49),
    ActDep('search furniture', 'you come across an old box'),
    LocDep('pocket valley', loc=53),
    EntDep(['pile of oats', 'oats', 'pile'], loc=53),
    ActDep('search oats', 'You find a shiny scroll!'),
    InvDep('shiny scroll', 62)
]
\end{lstlisting}

To satisfy the \textbf{Entity Dependency} (EntDep), the agent must detect an Entity called "furniture" or "wooden furniture" at Location 49. This dependency is verified by using Jericho's introspection feature to detect when the agent visits the location corresponding to world object number 49, and looking into the Knowledge Graph for an Entity named furniture at the KG's current Location. The \textbf{Action Dependency} (ActDep) "search furniture" is satisfied upon receiving a new observation that contains the text "you come across an old box." This is verified simply by monitoring the incoming observations through the event stream. Simple text matching is sufficient to recognize the results of key actions that progress the game. The \textbf{Location Dependency} (LocDep) is satisfied when the agent visits location number 53 (aka Pocket Valley). Locations are verified by using Jericho to inspect the game and return the world object number corresponding to the player's actual location. Finally the \textbf{Inventory Dependency} (InvDep) is satisfied when the world object number 62 exists in the player's inventory. This is verified using Jericho to access the list of world objects belonging to the player.

\begin{figure}[htp]
\centering
\includegraphics[width=.5\linewidth]{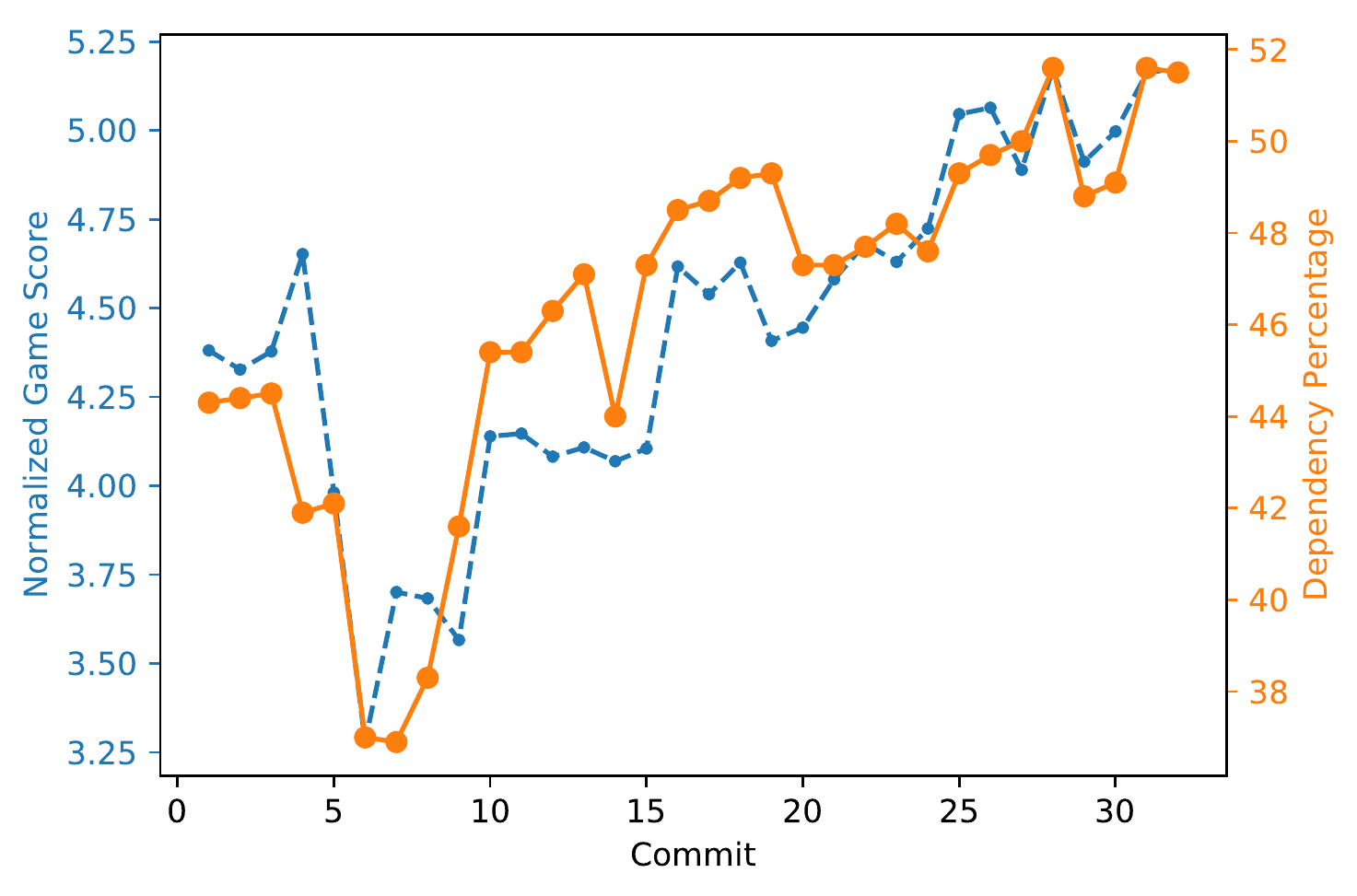}
\caption{\textbf{Development of NAIL on fifty-six human-made IF games}: The blue line tracks normalized game score while the orange shows the percentage of dependencies that are satisfied. The x-axis tracks the 32 commits made to NAIL after the implementation of game-specific Dependencies.}
\label{fig:deps}
\end{figure}

We implemented Dependencies for each of the fifty-six games in the Jericho suite. Figure \ref{fig:deps} shows that these Dependencies help quantify progress and more importantly, help pinpoint exactly which parts of the agent need to be improved: if the agent is failing ActDeps, then perhaps the Interactor needs to use a different set of verbs, conversely if NavDeps are failing, a bug may have entered the Navigator.

\section{Results}
NAIL won first place in the 2018 Text-Based Adventure AI Competition~\cite{atkinson18}, where it was evaluated on a set of twenty unknown parser-based IF games. Designed to assess an agent's ability to achieve human-like performance, the competition only allowed one thousand steps of interaction per game, comparable to a few hours of playtime for a human. Each agent's scores were normalized by the maximum possible score per game, then averaged over all games to obtain the final scores shown in Table \ref{tab:competition_results}.


\begin{table}[htp]
\centering
\begin{tabular}{@{}lrrrrr@{}}
\toprule
Agent & \multicolumn{2}{c}{\% completion} & \multicolumn{2}{c}{\% non-zero} \\
\cmidrule{2-3} \cmidrule{4-5}
& M & SD & M & SD \\
\midrule
\textsc{BYU\-Agent 2016} & 0.79 & - & 15 & - \\
\textsc{Golovin} & 1.45 & 0.09 & 31 & 3.94 \\
CARL (\textsc{BYU\-Agent 2017}) & 1.59 & - & 30 & - \\
NAIL & {\bf 2.56} & 0.33 & {\bf 45.5} & 2.84 \\
\midrule
\textsc{Golovin} (100 steps) & 0.99 & 0.24 & 17.5 & 3.53 \\
NAIL (100 steps) & 0.95 & 0.19 & 26 & 2.11 \\
\textsc{Golovin} (10k steps) & 1.44 & 0.10 & 32.5 & 4.25 \\
\midrule
\textit{RandomAgent} & 1.66 & 0.15 & 34 & 2.11 \\
\bottomrule
\end{tabular}
\vspace{1em}
\caption{Performance on the test set of 20 games in (unless stated otherwise) 1000 time steps per game. ``\% completion'' is the average score percentage an agent achieved over all games and runs; ``\% non-zero'' is the percentage of games in which an agent achieved any score, averaged over all runs. Standard deviations (SD), wherever given, refer to 10 runs over all games. Where they are not given, only 1 run could be completed. Table duplicated from \cite{atkinson18}.}
\label{tab:competition_results}
\end{table}

\section{Analysis}
As apparent from the competition results, agents have a long way to go towards solving unseen games. However, the progress over the past three years of the competition is encouraging. NAIL advances the state-of-the-art in comparison to other agents in several ways: 

\begin{enumerate}
    \item NAIL maintains an explicit Knowledge Graph which tracks relevant game information and builds a map of the game world. The information contained within is both used and populated by decision modules as the game progresses. This knowledge representation is human-interpretable and debuggable given ground-truth information about the game.
    \item Unlike prior agents, NAIL leverages the intuition that interactive objects can be examined, and extensively uses its Examiner decision module as a gatekeeper for deciding which objects are worth interacting with. Across the training set of $56$ games, $26\%$ of NAIL's actions are Examines, versus $8\%$ for CARL, $2\%$ for Golovin, and only $0.2\%$ for BYU. By exhaustively examining candidate objects, NAIL can focus actions on only the objects that are recognized by the game's parser.
    \item NAIL is the first agent to use a Validity Detector, a learned model, to decide whether actions have succeeded. This model is key to correctly populating the Knowledge Graph and is used extensively by individual decision modules to reason about the success of their actions.
\end{enumerate}

\begin{figure}[htp]
\centering
\includegraphics[width=.4\linewidth]{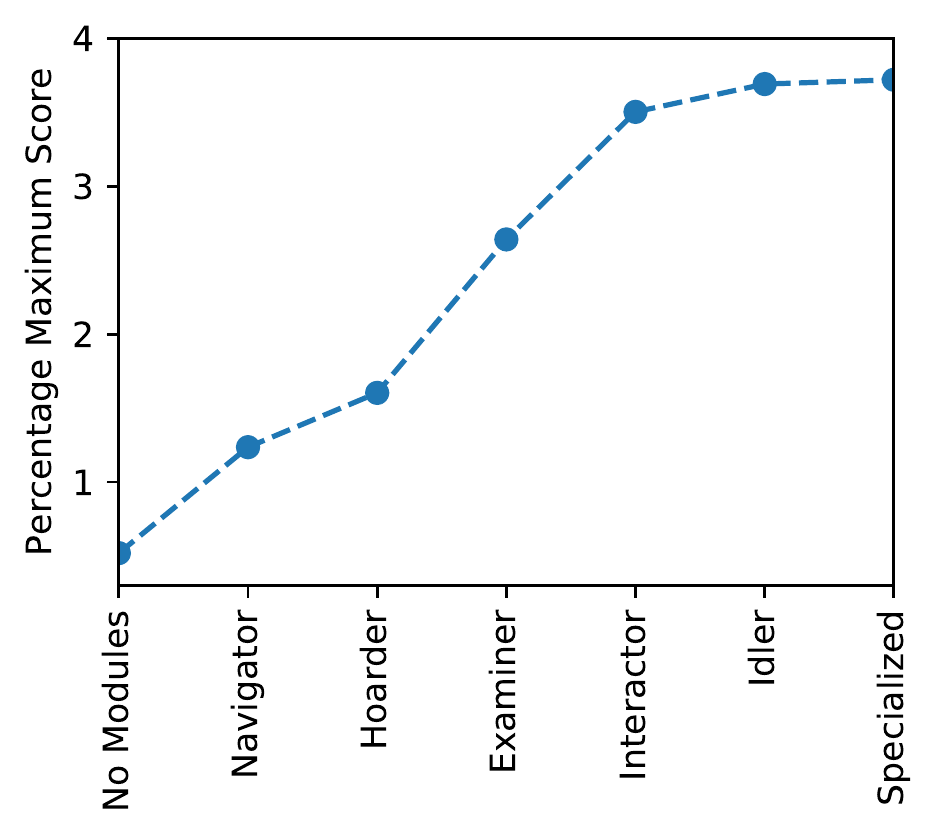}
\caption{\textbf{Ablation of decision modules:} Largest performance increases come from NAIL's core decision modules: the Navigator, the Examiner, and the Interactor.}
\label{fig:ablation}
\end{figure}

To further understand NAIL's performance, we selectively ablate NAIL's decision modules. Figure~\ref{fig:ablation} shows the average normalized score of NAIL across Jericho games as decision modules are sequentially added. Without any decision modules, NAIL is capable of only performing the "look" command and gets a score of $0.53\%$, since some games start with small positive score. Adding the Navigator allows the agent to locomote and create a map of the game world. Using this module, scores go up to $1.2\%$, primarily due to games that reward visiting new locations. Next, the Hoarder issues "take all" commands at each new location it visits to collect items. Scores increase to $1.6\%$ since many games reward the player for acquiring treasures or key items. Adding the Examiner module allows the agent to more deeply search the environment and reason about which objects are interactive, boosting the score to $2.6\%$. Leveraging the objects identified by the Examiner, the Interactor uses its language model to generate likely actions for application to those objects. These interactions are key to solving puzzles and boost the score to $3.5\%$. Together, these modules make up NAIL's core and account for the lion's share of the score. A $0.2\%$ gain is added by the Idler which exhaustively generates common IF actions when no other module is eager to take control. Finally, the specialized decision modules together contribute $0.03\%$.

\section{Discussion}
Interactive Fiction games are rich narrative adventures that challenge even skilled human players. We presented NAIL, an open-source,\footnote{NAIL's source code  is available at~\href{https://github.com/Microsoft/nail_agent}{https://github.com/Microsoft/nail\_agent}.} competition-winning IF agent. More than just a baseline for future comparison, we expect that NAIL's extendable architecture can serve as a starting point for future IF agents.

\section{Acknowledgements}
The authors would like to thank Marc-Alexandre C\^ot\'e, Xingdi Yuan, and Alekh Agarwal for their comments and suggestions. Additional thanks to Shuohang Wang for testing models for learning priorities over examined objects.

\bibliography{main}
\bibliographystyle{plainnat}
\appendix

\begin{table}[htp]
    \centering
    \begin{scriptsize}
    \begin{tabular}{lrrrrrr}
        \textbf{Game} & \textbf{Random} & \textbf{BYU `16} & \textbf{Golovin} & \textbf{CARL} & \textbf{NAIL}\\
        905 & 0.0 & 0.0 & 0.0 & 0.0 & 0.0\\
        acorncourt & 0.0 & 0.0 & 0.0 & 0.0 & 0.0\\
        advent & 36.0 & 36.0 & 36.0 & 36.0 & 36.0\\
        adventureland & 0.0 & 0.0 & 0.0 & 0.0 & 0.0\\
        afflicted & 0.0 & 0.0 & 0.0 & 0.0 & 0.0\\
        anchor & 0.0 & 0.0 & 0.0 & 0.0 & 0.0\\
        awaken & 0.0 & 0.0 & 0.0 & 0.0 & 0.0\\
        balances & 0.0 & 0.9 & \textbf{10.0} & 9.1 & \textbf{10.0}\\
        ballyhoo & 0.0 & 0.0 & \textbf{0.6} & \textbf{0.6} & 0.0\\
        curses & 0.0 & 0.0 & 0.2 & 0.0 & \textbf{1.7}\\
        cutthroat & 0.0 & 0.0 & 0.0 & 0.0 & 0.0\\
        deephome & 1.0 & 1.0 & 1.0 & 2.2 & \textbf{13.3}\\
        detective & 113.7 & 22.5 & 56.2 & 58.1 & \textbf{136.9}\\
        dragon & 0.0 & 0.0 & \textbf{0.8} & 0.0 & 0.6\\
        enchanter & 0.0 & 0.0 & 0.0 & 0.0 & 0.0\\
        enter & 0.0 & 0.0 & \textbf{0.4} & 0.0 & 0.0\\
        gold & 0.0 & 0.0 & 0.0 & 0.0 & \textbf{3.0}\\
        hhgg & 0.0 & 0.0 & 0.0 & 0.0 & 0.0\\
        hollywood & 0.0 & 0.0 & 0.0 & 0.0 & 0.0\\
        huntdark & 0.0 & 0.0 & 0.0 & 0.0 & 0.0\\
        infidel & 0.0 & 0.0 & 0.0 & 0.0 & 0.0\\
        inhumane & 0.0 & 0.0 & 0.0 & 0.0 & \textbf{0.6}\\
        jewel & 0.0 & 0.0 & 0.0 & 0.0 & \textbf{1.6}\\
        karn & 0.0 & 0.0 & 0.0 & 0.0 & \textbf{1.2}\\
        lgop & 0.0 & 0.0 & 0.0 & 5.5 & \textbf{6.1}\\
        library & 0.0 & 0.6 & \textbf{5.1} & 0.0 & 0.9\\
        loose & 0.0 & 0.0 & 0.0 & \textbf{1.1} & 0.0\\
        lostpig & 1.0 & 1.0 & 1.2 & 1.4 & \textbf{2.0}\\
        ludicorp & \textbf{13.2} & 2.8 & 1.0 & 8.3 & 8.4\\
        lurking & 0.0 & 0.0 & 0.0 & 0.0 & \textbf{0.6}\\
        moonlit & 0.0 & 0.0 & 0.0 & 0.0 & 0.0 \\
        murdac & 13.7 & 4.8 & 13.6 & \textbf{14.0} & 13.8\\
        omniquest & 0.0 & 5.0 & \textbf{5.6} & 5.0 & \textbf{5.6}\\
        partyfoul & 0.0 & 0.0 & 0.0 & 0.0 & 0.0\\
        pentari & 0.0 & 0.0 & 0.0 & 0.0 & 0.0\\
        planetfall & 0.2 & 0.0 & 0.0 & 0.2 & \textbf{0.6}\\
        plundered & 0.0 & 0.0 & 0.0 & 0.0 & 0.0\\
        reverb & 0.0 & 0.0 & 0.0 & 0.0 & 0.0\\
        seastalker & 1.0 & 0.1 & 0.3 & \textbf{2.4} & 1.6\\
        sherlock & 0.0 & 0.6 & 0.0 & \textbf{0.8} & 0.0\\
        snacktime & 0.0 & 0.0 & 0.0 & 0.0 & 0.0\\
        sorcerer & 5.0 & 5.0 & 5.0 & 5.0 & 5.0\\
        spellbrkr & 25.0 & 22.8 & \textbf{40.0} & \textbf{40.0} & \textbf{40.0}\\
        spirit & \textbf{2.4} & 0.0 & 0.5 & 1.1 & 1.0\\
        temple & 0.0 & 0.0 & 0.0 & 0.3 & \textbf{7.3}\\
        theatre & 0.0 & 0.0 & 0.0 & 0.0 & 0.0\\
        trinity & \textbf{3.9} & 0.2 & 1.6 & 3.0 & 2.6\\
        tryst205 & 0.0 & 0.0 & 0.7 & 0.0 & \textbf{2.0}\\
        weapon & 0.0 & 0.0 & 0.0 & 0.0 & 0.0\\
        wishbringer & \textbf{6.4} & 3.4 & 1.6 & 6.0 & 6.0\\
        yomomma & 0.0 & 0.0 & 0.0 & 0.0 & 0.0\\
        zenon & 0.0 & 0.0 & 0.0 & 0.0 & 0.0\\
        zork1 & 0.0 & 2.5 & 7.2 & 9.1 & \textbf{10.3}\\
        zork2 & 0.0 & 0.0 & 0.0 & 0.0 & 0.0\\
        zork3 & 0.2 & 0.0 & 1.1 & 0.3 & \textbf{1.8}\\
        ztuu & 0.0 & \textbf{3.4} & 0.0 & 0.0 & 0.0\\
        \textbf{Normalized Score} & 1.68 & 1.11 & 2.37 & 2.16 & \textbf{3.72} \\
        \textbf{Times Best} & 4 & 1 & 3 & 4 & \textbf{15} \\
	\end{tabular}
    \end{scriptsize}
    \caption{\label{tab:scores-full} \textbf{Raw scores} for Jericho-supported games, averaged over sixteen runs. The Random agent selects actions from the set \texttt{north/south/east/west/up/down/look/inventory/take all/drop/yes}.}
\end{table}

\newpage
\section{Analysis of Parser-Based IF Games}
\label{sec:appendix_analysis}
Perhaps the least friendly user interface of all time, parser-based IF games accept any natural language string as input, and use a parser to interpret the player's action. The difficulty of using the interface stems from the fact that many natural language strings are not recognized by the parser and result in failed actions. For example, many games will produce canned  responses such as ``I don't know the word \textit{x}.'' or ``You can't \textit{y}.'' Since the parser is hidden, players often need to read a manual and experiment with the game to discover what types of actions are recognized. Fortunately, parsers for many popular IF games are similar in the types of actions they accept and the responses they produce for unrecognized actions. This standardization reduces the burden on learning agents, as they do not have to generate arbitrarily complex natural language.

\begin{figure}[t]
\centering
\begin{subfigure}{.4\linewidth}
  \centering
  \includegraphics[width=\linewidth]{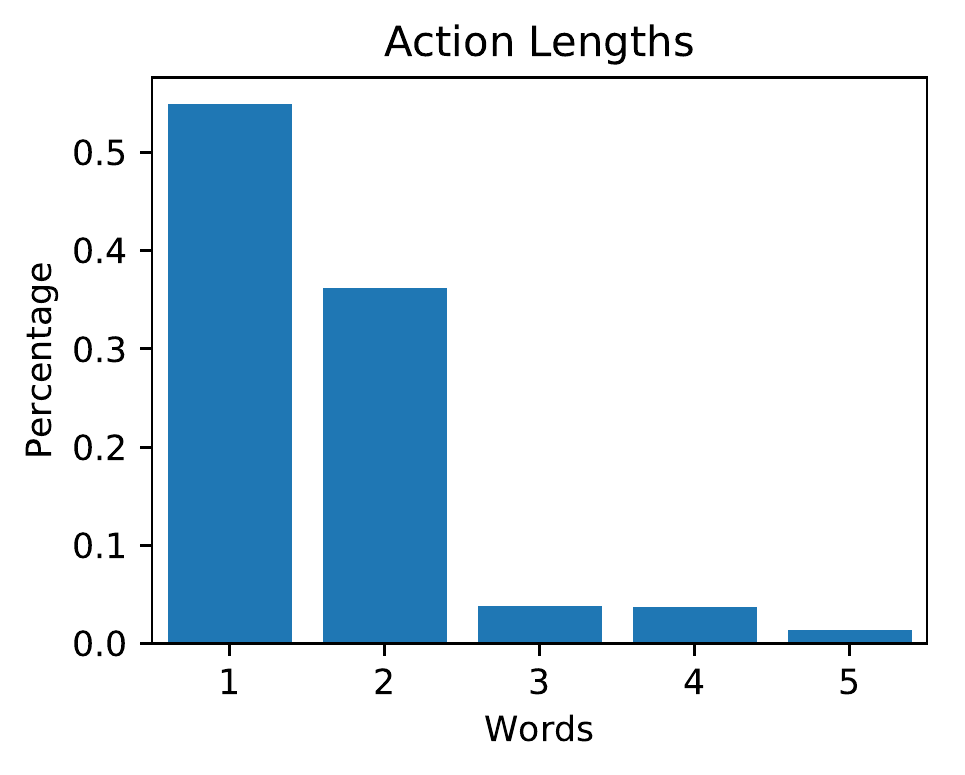}
\end{subfigure}%
\begin{subfigure}{.4\linewidth}
  \centering
  \includegraphics[width=\linewidth]{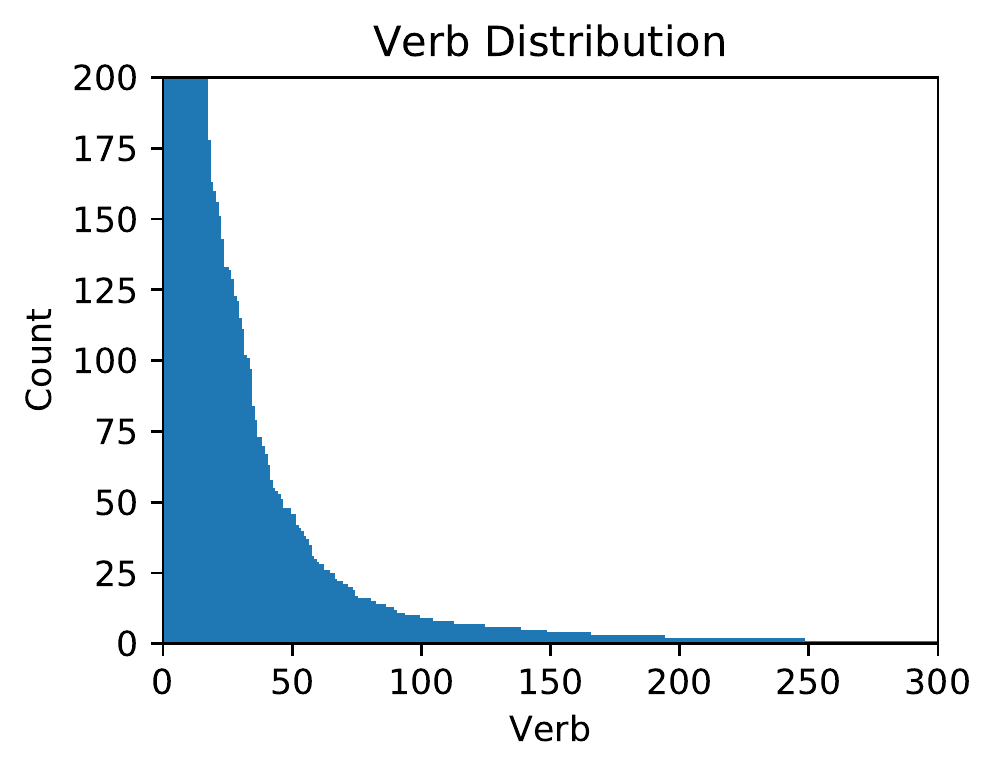}
\end{subfigure}
\caption{\textbf{Analysis of walkthroughs} reveals that over 90\% of actions are one and two words in length. Among these actions, there were 530 unique verbs used, but the 100 most common account for 95\% of all actions.}
\label{fig:action_hist}
\end{figure}

To better understand the complex action space of parser-based games, we analyzed human-created walkthroughs for 188 games. From these walkthroughs we extracted 20,263 natural language actions. As shown in Figure \ref{fig:action_hist} (left), most actions are one or two words in length, with a maximum of five words\footnote{All actions with six words or longer were reducible to equivalent shorter actions.}. Further analysis reveals that these actions have extensive structure: single-word actions are often shortcuts provided by the game for navigation (``north'' moves the player north), examination (``look'' describes the current location), and item management (``inventory'' lists the objects carried). Two-word actions take the form of \textit{verb-object} (``climb tree'', ``eat apple''). Three-word actions are commonly \textit{verb-preposition-object} (``search under bed''), but can occasionally take on different patterns (``turn dial left''). Though uncommon, four-word actions include the pattern \textit{verb-object-preposition-object} (``unlock chest with key'', ``ask bolitho about ghost''). Five-word actions commonly used multiple words to describe an object: ``attack troll with brass lantern.''

The verb distribution shows that the majority of actions stem from a compact set of verbs focused on navigation, item acquisition, and environment examination. However, the distribution in Figure \ref{fig:action_hist} (right) has a long tail corresponding to a diverse set of verbs used to interact with objects in the environment. The nouns used in these commands are highly varied from game to game. Some games even go so far as to create their own proper nouns for special objects and spells. Such words are not in any English dictionary and need to be remembered from the observation text.

Altogether, this analysis indicates that the action generation task in IF games is significantly more structured than generating free-form dialog. Thus, while learning agents still need to \textit{understand} arbitrary free-form text presented by the game, they only need to \textit{generate} a compact subset of language. 

\end{document}